\pdfoutput=1

\documentclass[11pt]{article}

\usepackage{ACL2023}
\usepackage{graphicx}
\usepackage{sans}
\usepackage{times}
\usepackage{latexsym}

\PassOptionsToClass{sort&compress,numbers}{natbib}

\usepackage[T1]{fontenc}

\usepackage[utf8]{inputenc}

\usepackage{microtype}

\usepackage{amsmath}

\usepackage{inconsolata}

\usepackage{xcolor}
\usepackage{xspace}
\usepackage{cleveref}
\usepackage{multicol}
\usepackage{multirow}
\usepackage{booktabs}
\usepackage{float}
\usepackage{comment}
\usepackage[normalem]{ulem}
\useunder{\uline}{\ul}{}
\usepackage{enumitem}

\newcommand{\cy}[1]{{\small\textcolor{orange}{\bf [#1 --CY]}}}

\newcommand{\eg}{{\it e.g.}\xspace}
\newcommand{\ie}{{\it i.e.}\xspace}

\title{Good Data, Large Data, or No Data? Comparing Three Approaches in Developing Research Aspect Classifiers for Biomedical Papers}

\author{Shreya Chandrasekhar, Chieh-Yang Huang, and Ting-Hao `Kenneth' Huang \\
  Pennsylvania State University, University Park, PA, USA \\
  \texttt{\{sxc6175, chiehyang, txh710\}@psu.edu} \\
}
  
\begin{document}

\maketitle
\begin{abstract}

The rapid growth of scientific publications, particularly during the COVID-19 pandemic, 
emphasizes the need for tools to help researchers efficiently comprehend the latest advancements. 
One essential part of understanding scientific literature is research aspect classification,
which categorizes sentences in abstracts to Background, Purpose, Method, and Finding.
In this study, we investigate the impact of different datasets on model performance for
the crowd-annotated CODA-19 research aspect classification task.
Specifically, we explore the potential benefits of using the large, automatically curated PubMed 200K RCT dataset 
and evaluate the effectiveness of large language models (LLMs), such as LLaMA, GPT-3, ChatGPT, and GPT-4.
Our results indicate that using the PubMed 200K RCT dataset does not improve performance for the CODA-19 task.
We also observe that while GPT-4 performs well,
it does not outperform the SciBERT model fine-tuned on the CODA-19 dataset,
emphasizing the importance of a dedicated and task-aligned datasets dataset for the target task.
Our code is available at \url{https://github.com/Crowd-AI-Lab/CODA-19-exp}.

\end{abstract}


\section{Introduction}

The rapid growth of scientific publications, particularly during the COVID-19 pandemic, 
has made it increasingly challenging to keep up with the latest research advancements.
To address this issue, researchers have developed various systems,
such as search engines~\cite{lahav2022search,zhang-etal-2020-rapidly},
visualization tools~\cite{hope2020scisight,tu2020exploration},
claim verification systems~\cite{wadden-etal-2020-fact,pradeep-etal-2021-scientific}, 
question-answering systems~\cite{frisoni-etal-2022-bioreader},
and summarization techniques~\cite{meng-etal-2021-bringing}.
These tools help efficiently comprehend publications by organizing large amounts of information.

One critical aspect of understanding scientific literature is the classification of sentences
within abstracts according to their research aspects, such as Background, Purpose, Method, and Finding.
This is particularly crucial in the biomedical domain, where abstracts tend to be longer and more complex.
With the annotated aspects, readers can quickly grasp the key aspects of a scientific paper.
For example, FacetSum~\cite{meng-etal-2021-bringing} summarizes papers in four aspects to quickly convey the information.
Several datasets have been proposed for research aspect classification,
including PubMed 200K RCT~\cite{dernoncourt2017pubmed}, PubMed-PICO-Detection~\cite{jin-szolovits-2018-pico},
CODA-19~\cite{huang2020coda}, and more.

In this paper, we focus on the \textbf{research aspect classification task using crowd-annotated CODA-19}~\cite{huang2020coda}
and explore the impact of different datasets on model performance.
Specifically, we investigate whether the \textbf{automatically curated large dataset} (PubMed 200K RCT~\cite{dernoncourt2017pubmed})
can help the target task despite its shifted data distribution from the target task.
Additionally, we examine whether large language models (LLMs), trained on a massive general textual corpus,
can solve the task with \textbf{limited or no task-specific data provided}.
In particular, we evaluate six LLMs, including three open-sourced models (LLaMA-65B~\cite{touvron2023llama}, MPT~\cite{MosaicML2023Introducing}, and Dolly-12B~\cite{dolly-12b}) 
and three closed models (GPT-3~\cite{brown2020language}, ChatGPT~\cite{chatgpt}, and GPT-4~\cite{openai2023gpt4}).

Our study suggests that using PubMed 200K \textbf{does not} improve the performance of 
the CODA-19 research aspect classification task, regardless of the approach used.
We experimented with (\textit{i}) training purely on PubMed data, 
(\textit{ii}) simply-mixing PubMed and CODA-19 data,
(\textit{iii}) upsampling CODA-19 data and mixing it with PubMed data to address the data imbalance issue, 
and (\textit{iv}) using a two-staged training approach where the model is trained on PubMed and CODA-19 sequentially.
However, none of these approaches could improve the performance of the target task.
We hypothesize this is due to the use of the SciBERT~\cite{beltagy-etal-2019-scibert},
which has pre-trained on papers from scientific domains and 
thus reduces the advantage of incorporating PubMed 200K.
Our results also showed that although GPT-4 performed well in both zero-shot and few-shot settings, 
\textbf{it was not able to outperform} the SciBERT model fine-tuned on the CODA-19 dataset.
This finding suggests that having a dedicated dataset that aligns well with the target task is still important.

\section{Related Work}






\citet{moradi2021gpt} compared BioBERT and GPT-3 in a few-shot learning setting for sentence classification tasks and found that neither model could outperform fully fine-tuned models.
For biomedical information extraction tasks, 
\citet{jimenez-gutierrez-etal-2022-thinking} observed that GPT-3 could not outperform the fine-tuned RoBERTa-large model, 
whereas \citet{agrawal-etal-2022-large} reported promising performance for GPT-3.
\citet{gururangan-etal-2020-dont} investigated whether adapting pre-trained models to the domain of the target task could help, and concluded that domain-adaptive pretraining is always helpful, 
highlighting the need for task-aligned data and training strategies.
In this paper, we focus on the aspect classification task on CODA-19 and examine whether slightly domain-shifted large datasets and LLMs can improve performance.

\section{Methodology}

In this section, we describe models trained to explore the impact of different datasets and training strategies on the CODA-19 research aspect classification task. For detailed model training details such as hyperparameters, please refer to \Cref{appendix:hyperparameters}.



\subsection{Good Data: CODA-19}
\label{sec:coda-model}
We use CODA-19~\cite{huang2020coda} as our Good Data.
CODA-19 consists of clause-level aspect annotations for abstracts from medical papers,
including \textit{Background, Purpose, Method, Finding/Contribution, and Other}.
It contains 137K/15K/15K samples for train/validation/test sets.
We fine-tune SciBERT~\cite{beltagy-etal-2019-scibert} on the original CODA-19 to create SciBERT\textsubscript{CODA} and on the position-encoded CODA-19 to create SciBERT\textsubscript{CODA+Pos} (see \Cref{sec:position_encoding}).

\subsection{Large Data: PubMed}
\label{sec:pubmed-model}
PubMed 200K~\cite{dernoncourt2017pubmed} provides sentence-level research aspects based on structured abstracts.
It contains 2.2M/29K/29K samples for the train/validation/test set.
Despite including a larger amount of data, the distribution is slightly different from the CODA-19 task.
We first create two models, SciBERT\textsubscript{PubMed} and SciBERT\textsubscript{PubMed+Pos}, by fine-tuning SciBERT on PubMed 200K.
We also explore combining CODA-19 and PubMed with three strategies:
(\textit{i}) simply-mixing the two datasets to create SciBERT\textsubscript{Mix+Pos+S};
(\textit{ii}) upsampling CODA-19 ten times to balance the data size and mixing it with PubMed to create SciBERT\textsubscript{Mix+Pos+U};
and (\textit{iii}) two-staged training, where the model is fine-tuned on PubMed and CODA-19 sequentially to create SciBERT\textsubscript{Mix+Pos+T}.

\begin{table*}[t]
\small \centering
\addtolength{\tabcolsep}{-1.1mm}
\begin{tabular}{@{}llccccccccccccccccc@{}}
\toprule
\multicolumn{2}{c}{\textbf{Setting}} & \multicolumn{3}{c}{\textbf{Background}} & \multicolumn{3}{c}{\textbf{Methods}} & \multicolumn{3}{c}{\textbf{Objective}} & \multicolumn{3}{c}{\textbf{Results}} & \multicolumn{3}{c}{\textbf{Conclusions}} & \multicolumn{2}{c}{\textbf{All}} \\ \midrule
\multicolumn{2}{c}{\textbf{Support}} & \multicolumn{3}{c}{\textbf{2,663}} & \multicolumn{3}{c}{\textbf{9,751}} & \multicolumn{3}{c}{\textbf{2,377}} & \multicolumn{3}{c}{\textbf{10,276}} & \multicolumn{3}{c}{\textbf{4,426}} & \multicolumn{2}{c}{\textbf{29,493}} \\ \midrule
\textbf{Model} & \textbf{FTR} & \textbf{P} & \textbf{R} & \textbf{F1} & \textbf{P} & \textbf{R} & \textbf{F1} & \textbf{P} & \textbf{R} & \textbf{F1} & \textbf{P} & \textbf{R} & \textbf{F1} & \textbf{P} & \textbf{R} & \textbf{F1} & \textbf{ACC} & \textbf{F1} \\ \midrule
\textbf{SciBERT} & \textbf{T} & {\ul .712} & .734 & .723 & .935 & .955 & .945 & {\ul .787} & {\ul .655} & {\ul .715} & .918 & .932 & .925 & .858 & .848 & .853 & .887 & .832 \\
\textbf{SciBERT} & \textbf{T+P} & \textbf{.742} & \textbf{.824} & \textbf{.781} & {\ul .945} & \textbf{.970} & {\ul .958} & \textbf{.796} & \textbf{.679} & \textbf{.733} & \textbf{.959} & {\ul .940} & {\ul .950} & {\ul .946} & \textbf{.946} & \textbf{.946} & \textbf{.919} & \textbf{.873} \\
\textbf{bi-ANN+CRF} & \textbf{T} & .707 & {\ul .811} & {\ul .756} & \textbf{.955} & {\ul .965} & \textbf{.960} & .771 & .653 & .707 & {\ul .956} & \textbf{.948} & \textbf{.952} & \textbf{.946} & {\ul .937} & {\ul .942} & {\ul .916} & {\ul .863} \\ \bottomrule
\end{tabular}
\addtolength{\tabcolsep}{+1.2mm}
\caption{Performance on PubMed 200k~\cite{dernoncourt2017pubmed}, where \textbf{best} and {\ul second-best} results are highlighted.
The FTR column shows the feature used for the model, where T and T+P stand for text-only and position-encoded text.
According to the overall accuracy and F1 score, our SciBERT\textsubscript{PubMed+Pos} performs the best.}
\label{tab:pubmed-exp}
\end{table*}


\begin{table*}[t]
\small \centering
\addtolength{\tabcolsep}{-1.1mm}
\begin{tabular}{@{}lllccccccccccccccccc@{}}
\toprule
\multicolumn{3}{c}{\textbf{Setting}} & \multicolumn{3}{c}{\textbf{Background}} & \multicolumn{3}{c}{\textbf{Purpose}} & \multicolumn{3}{c}{\textbf{Method}} & \multicolumn{3}{c}{\textbf{Finding}} & \multicolumn{3}{c}{\textbf{Other}} & \multicolumn{2}{c}{\textbf{ALL}} \\ \midrule
\multicolumn{3}{c}{\textbf{Support}} & \multicolumn{3}{c}{\textbf{5,062}} & \multicolumn{3}{c}{\textbf{821}} & \multicolumn{3}{c}{\textbf{2,140}} & \multicolumn{3}{c}{\textbf{6,890}} & \multicolumn{3}{c}{\textbf{562}} & \multicolumn{2}{c}{\textbf{15,475}} \\ \midrule
\textbf{Data} & \textbf{Model} & \textbf{FTR} & \textbf{P} & \textbf{R} & \textbf{F1} & \textbf{P} & \textbf{R} & \textbf{F1} & \textbf{P} & \textbf{R} & \textbf{F1} & \textbf{P} & \textbf{R} & \textbf{F1} & \textbf{P} & \textbf{R} & \textbf{F1} & \textbf{ACC} & \textbf{F1} \\ \midrule

\textbf{C} & \textbf{SciBERT} & \textbf{T} & .700 & .808 & .750 & .656 & {\ul .588} & .620 & .716 & .635 & .673 & .802 & .743 & .772 & .797 & \textbf{.867} & .830 & .746 & .729 \\
\textbf{C} & \textbf{SciBERT} & \textbf{T+P} & {\ul .825} & .794 & .809 & .638 & \textbf{.655} & \textbf{.647} & \textbf{.741} & .665 & \textbf{.701} & .823 & {\ul .867} & {\ul .845} & \textbf{.819} & .843 & {\ul .831} & {\ul .803} & \textbf{.767} \\ \midrule

\textbf{P} & \textbf{SciBERT} & \textbf{T} & .710 & .433 & .538 & .415 & .167 & .238 & .371 & \textbf{.757} & .498 & .678 & .756 & .715 & - & - & - & .592 & .497 \\
\textbf{P} & \textbf{SciBERT} & \textbf{T+P} & \textbf{.854} & .415 & .559 & .233 & .251 & .241 & .362 & {\ul .720} & .482 & .750 & .857 & .800 & - & - & - & .630 & .520 \\ \midrule

\textbf{Mix}$_\text{s}$ & \textbf{SciBERT} & \textbf{T+P} & .762 & {\ul .815} & .788 & {\ul .674} & .446 & .537 & .677 & .640 & .658 & .824 & .843 & .833 & .808 & .642 & .716 & .777 & .706 \\
\textbf{Mix}$_\text{u}$ & \textbf{SciBERT} & \textbf{T+P} & .802 & \textbf{.822} & {\ul .812} & .669 & .585 & .624 & .734 & .636 & .681 & {\ul .826} & .857 & .841 & {\ul .812} & .820 & .816 & .799 & .755 \\
\textbf{Mix}$_\text{t}$ & \textbf{SciBERT} & \textbf{T+P} & .816 & .812 & \textbf{.814} & \textbf{.687} & .574 & {\ul .625} & {\ul .736} & .654 & {\ul .693} & \textbf{.827} & \textbf{.870} & \textbf{.848} & .807 & {\ul .865} & \textbf{.835} & \textbf{.805} & {\ul .763} \\ \midrule

\textbf{C} & \textbf{BERT} & \textbf{T+P} & {\ul .846} & .761 & .801 & .626 & {\ul .646} & .636 & .702 & .637 & .668 & .803 & \textbf{.879} & .839 & .803 & .847 & .824 & .793 & .754 \\
\textbf{Mix}$_\text{t}$ & \textbf{BERT} & \textbf{T+P} & .828 & .775 & .801 & .663 & .639 & \textbf{.651} & .715 & .639 & .675 & .808 & {\ul .872} & .839 & .809 & .854 & {\ul .831} & .795 & .759 \\

\bottomrule
\end{tabular}
\addtolength{\tabcolsep}{+1.1mm}
\caption{Performance on CODA-19~\cite{huang2020coda}, where \textbf{best} and {\ul second-best} results are highlighted. The Data column specifies the training data, C: CODA-19, P: PubMed 200K, Mix\textsubscript{s}: simply-mixing, Mix\textsubscript{u}: upsampling, and Mix\textsubscript{t}: two-staged training. The FTR column shows the feature used for the model, T: text-only, and T+P: position-encoded text. According to the overall performance,
 SciBERT\textsubscript{CODA+Pos} and SciBERT\textsubscript{Mix+Pos+T} achieve the best performance.}
\label{tab:exp-coda-19}
\end{table*}

\subsection{No Data: LLMs}
In the No-Data setting, the goal is to classify the research aspect of abstract sentences using LLMs with limited or no task-specific training data.
LLMs are trained on a massive amount of web data, which has a different distribution compared to our target dataset. 
To explore the performance of LLMs in this scenario, we use zero-shot and few-shot classification with 
three open-sourced models: LLaMA-65B~\cite{touvron2023llama}, MPT-7B~\cite{MosaicML2023Introducing}, and Dolly-12B~\cite{dolly-12b};
and three closed models: GPT-3~\cite{brown2020language}, ChatGPT~\cite{chatgpt}, and GPT-4~\cite{openai2023gpt4}.

\subsection{Position Encoding}
\label{sec:position_encoding}
To predict a research aspect for a sentence, the position of the sentence within the whole abstract is important.
Prior work such as \citet{dernoncourt2017pubmed} used CRF to model the relationship between sentences.
In this paper, we incorporate position information by simply adding a position encoding to the beginning of each sentence in the form of ``[POSITION=0.38]'', where the number represents the normalized sentence position, i.e., sentence\_id / \#sentences\_in\_abstract.
Examples of position-encoded data can be found in \Cref{appendix:sample-data}.

\section{Experiments and Results}

We conducted three experiments to
(\textit{i}) verify whether our fine-tuned SciBERT model can outperform the model proposed in the original PubMed paper~\cite{dernoncourt2017pubmed};
(\textit{ii}) compare models trained on good data and large data, aiming to examine whether a large automatically curated dataset can enhance performance;
(\textit{iii}) benchmark the performance of open-sourced and closed LLMs for the CODA-19 aspect classification task and compare them with the best-performing SciBERT model.

\subsection{Verifying the PubMed Model}
In this experiment, we aim to assess the effectiveness of our fine-tuned PubMed model by comparing it with the model reported in the original PubMed paper~\cite{dernoncourt2017pubmed}.


\paragraph{Experimental Setup.}
We evaluate two PubMed models in our study: SciBERT\textsubscript{PubMed} and SciBERT\textsubscript{PubMed+Pos}.
To compare with the reported model, we apply them to PubMed 200K test set, which contains 29,493 samples, to predict the PubMed label set (\textit{Background, Methods, Objective, Results, and Conclusions}).
We report precision, recall, and F1 scores for each label and calculate the accuracy and macro F1 as overall metrics.
Note that the micro F1 score provided in the original PubMed paper is equivalent to accuracy since
each instance is assigned with only one label.
To obtain the macro F1 score, we average the F1 scores across all labels.


\paragraph{Results.}
The results shown in \Cref{tab:pubmed-exp} demonstrate that our SciBERT\textsubscript{PubMed+Pos} model outperforms the bi-ANN+CRF model~\cite{dernoncourt2017pubmed} in both of accuracy (0.919 vs. 0.916) and macro F1 score (0.873 vs. 0.863).
These findings suggest that, despite not considering a whole abstract simultaneously, we can achieve competitive performance by incorporating position encoding.
Based on these results, we use SciBERT\textsubscript{PubMed+Pos} for further comparisons.



\begin{table*}[]
\small \centering
\addtolength{\tabcolsep}{-1.25mm}
\begin{tabular}{@{}lllccccccccccccccccc@{}}
\toprule
\multicolumn{3}{c}{\textbf{Setting}} & \multicolumn{3}{c}{\textbf{Background}} & \multicolumn{3}{c}{\textbf{Purpose}} & \multicolumn{3}{c}{\textbf{Method}} & \multicolumn{3}{c}{\textbf{Finding}} & \multicolumn{3}{c}{\textbf{Other}} & \multicolumn{2}{c}{\textbf{ALL}} \\ \midrule
\multicolumn{3}{c}{\textbf{Support}} & \multicolumn{3}{c}{\textbf{250}} & \multicolumn{3}{c}{\textbf{250}} & \multicolumn{3}{c}{\textbf{250}} & \multicolumn{3}{c}{\textbf{250}} & \multicolumn{3}{c}{\textbf{250}} & \multicolumn{2}{c}{\textbf{1,250}} \\ \midrule
\textbf{Data} & \textbf{Model} & \textbf{FTR} & \textbf{P} & \textbf{R} & \textbf{F1} & \textbf{P} & \textbf{R} & \textbf{F1} & \textbf{P} & \textbf{R} & \textbf{F1} & \textbf{P} & \textbf{R} & \textbf{F1} & \textbf{P} & \textbf{R} & \textbf{F1} & \textbf{ACC} & \textbf{F1} \\ \midrule
\textbf{C} & \textbf{SciBERT} & \textbf{T+P} & .492 & .784 & \textbf{.605} & .879 & \textbf{.696} & \textbf{.777} & .710 & .684 & \textbf{.697} & .669 & .632 & \textbf{.650} & \textbf{.983} & .696 & \textbf{.815} & \textbf{.698} & \textbf{.709} \\
\textbf{P} & \textbf{SciBERT} & \textbf{T+P} & .241 & .080 & .120 & .633 & .152 & .245 & .288 & \textbf{.836} & .428 & .441 & .672 & .532 & - & - & - & .348 & .332 \\ \midrule
\textbf{-} & \multirow{2}{*}{\textbf{LLaMA}} & \textbf{Zero} & .212 & .160 & .182 & \textbf{1.000} & .012 & .024 & \textbf{1.000} & .004 & .008 & .700 & .056 & .104 & .180 & .748 & .291 & .196 & .122 \\
\textbf{-} &  & \textbf{Few} & .402 & .556 & .466 & .800 & .240 & .369 & .663 & .228 & .339 & .484 & .596 & .534 & .556 & .968 & .707 & .518 & .483 \\
\textbf{-} & \multirow{2}{*}{\textbf{MPT}} & \textbf{Zero} & .229 & \textbf{.960} & .370 & .000 & .000 & .000 & {\ul .923} & .048 & .091 & \textbf{1.000} & .004 & .008 & .746 & .564 & .642 & .315 & .222 \\
\textbf{-} &  & \textbf{Few} & .230 & .304 & .262 & \textbf{1.000} & .008 & .016 & .667 & .008 & .016 & .289 & {\ul .824} & .428 & {\ul .748} & .604 & .668 & .350 & .278 \\
\textbf{-} & \multirow{2}{*}{\textbf{Dolly}} & \textbf{Zero} & .208 & {\ul .956} & .342 & .000 & .000 & .000 & .000 & .000 & .000 & .304 & .096 & .146 & .522 & .048 & .088 & .220 & .115 \\
\textbf{-} &  & \textbf{Few} & .462 & .048 & .087 & .615 & .032 & .061 & .000 & .000 & .000 & .230 & \textbf{.904} & .367 & .652 & .592 & .621 & .315 & .227 \\ \midrule
\textbf{-} & \multirow{2}{*}{\textbf{GPT-3}} & \textbf{Zero} & .435 & .628 & .514 & .838 & .268 & .406 & .562 & .580 & .571 & .770 & .348 & .479 & .543 & .952 & .692 & .555 & .532 \\
\textbf{-} &  & \textbf{Few} & \textbf{.604} & .408 & .487 & .691 & .492 & .575 & .783 & .404 & .533 & .623 & .528 & .571 & .443 & \textbf{.996} & .613 & .566 & .556 \\
\textbf{-} & \multirow{2}{*}{\textbf{ChatGPT}} & \textbf{Zero} & .409 & .416 & .413 & .833 & .200 & .323 & .661 & .436 & .525 & .645 & .392 & .488 & .401 & {\ul .992} & .571 & .487 & .464 \\
\textbf{-} &  & \textbf{Few} & .446 & .516 & .479 & .833 & .200 & .323 & .663 & .464 & .546 & .621 & .472 & .536 & .461 & .988 & .628 & .528 & .502 \\
\textbf{-} & \multirow{2}{*}{\textbf{GPT-4}} & \textbf{Zero} & {\ul .579} & .560 & .569 & .749 & {\ul .548} & {\ul .633} & .562 & {\ul .692} & .620 & {\ul .800} & .416 & .547 & .615 & .952 & .747 & .634 & .623 \\
\textbf{-} &  & \textbf{Few} & .570 & .588 & {\ul .579} & {\ul .888} & .444 & .592 & .630 & .668 & {\ul .649} & .679 & .600 & {\ul .637} & .646 & .984 & {\ul .780} & {\ul .657} & {\ul .647} \\ \bottomrule
\end{tabular}

\addtolength{\tabcolsep}{+1.25mm}
\caption{Performance on a randomly sampled subset of CODA-19~\cite{huang2020coda}. We highlight the \textbf{best} and the {\ul second-best} results. The Data column specifies the training data: C: CODA-19 and P: PubMed.
The FTR column shows the feature used for the model, T: text-only, T+P: position-encoded text, zero: zero-shot learning, and few: few-shot learning.
Even the best-performing LLM, GPT-4, does not outperform SciBERT\textsubscript{CODA+Pos}, showing the need for task-aligned data. Open-sourced models (LLaMA, MPT, and Dolly) currently show lower performance compared to closed models.}
\label{tab:exp-coda-19-no-data}

\end{table*}

\subsection{Good Data and Large Data}
In this experiment, we aim to compare whether using a dataset with a larger amount of samples but a slight domain shift could help improve the performance of the CODA-19 aspect classification task.


\paragraph{Experimental Setup.}
\label{experiment}
We evaluate seven different models on CODA-19 test set, 
which consists of 15,475 samples across five labels: \textit{Background, Purpose, Method, Finding, and Other}.
The models include two trained on the CODA-19 dataset
(SciBERT\textsubscript{CODA} and SciBERT\textsubscript{CODA+Pos}), 
two trained on the PubMed dataset
(SciBERT\textsubscript{PubMed} and SciBERT\textsubscript{PubMed+Pos}),
and three trained on a mix of the CODA-19 and PubMed datasets
(SciBERT\textsubscript{Mix+Pos+S}, SciBERT\textsubscript{Mix+Pos+U}, and SciBERT\textsubscript{Mix+Pos+T}).
For the PubMed models, we map the predicted labels to the corresponding CODA-19 label using a predefined mapping function: 
\textit{Background} to \textit{Background}, \textit{Methods} to \textit{Method}, \textit{Objective} to \textit{Purpose}, \textit{Results} to \textit{Finding}, and \textit{Conclusions} to \textit{Finding}.
Note that \textit{Other} (unclear or confusing sentences) in CODA-19 does not have a corresponding label in PubMed.
We report precision, recall, and F1 scores for each label and calculate the accuracy and macro F1 as overall metrics.
For the PubMed models, the macro F1 score is averaged over the four valid labels.



\paragraph{Results.}
The results of our experiment are presented in \Cref{tab:exp-coda-19}.
We find that SciBERT\textsubscript{CODA+Pos} and SciBERT\textsubscript{Mix+Pos+T} achieve the highest accuracy (0.803 and 0.805) and macro F1 (0.767 and 0.763) scores, respectively, and outperform the other models.
The best performing PubMed model (SciBERT\textsubscript{PubMed+Pos}) does not show any improvement over the performance on the CODA-19 test set (accuracy: 0.630 and macro F1: 0.520); 
and is particularly weak in identifying the \textit{Purpose} label (\textit{Purpose} F1: 0.241).
When comparing the different mixing strategies, the models trained with simply-mixing and upsampling perform even worse (accuracy: 0.777/0.799 and macro F1: 0.706/0.755).
Although the two-staged mixing strategy does not yield lower scores, it only achieves the same results as SciBERT\textsubscript{CODA+Pos}.

Since SciBERT is pre-trained on a huge amount of scientific papers, with 82\% of the papers belonging to the biomedical domain \cite{beltagy-etal-2019-scibert}, we also compare the performance of two approaches, the two-staged mixing strategy, and pure fine-tuning, using BERT \cite{devlin-etal-2019-bert}.
This allows us to eliminate the impact of SciBERT's pretraining.
The results, shown in the last two rows of \Cref{tab:exp-coda-19}, indicate that the two-staged mixing strategy does not yield the expected improvement (with overall F1 scores of 0.754 for BERT\textsubscript{CODA+Pos} and 0.759 for BERT\textsubscript{Mix+Pos+T}).
Despite this, the overall score for SciBERT remains higher.
We hypothesize that the two-staged mixing strategy with the classification objective may not be the best way for adapting the model to a specific domain.
Therefore, when large-scale pretraining, such as training SciBERT, is not available, having a dedicated dataset that aligns well with the target task is still important and will give the best performance.

\subsection{Comparison of No-Data}
We investigate whether recent advances in LLMs can solve the CODA-19 aspect classification task.

\paragraph{Experimental Setup.}
Due to resource limitations, 
we experiment on a subset of CODA-19 test set, where 250 samples for each of the five labels
are randomly selected, resulting in 1,250 samples.

We first include both SciBERT\textsubscript{CODA+Pos} and SciBERT\textsubscript{PubMed+Pos} for comparison and report scores obtained by running on the evaluation set specifically for this experiment. 
Additionally, we include six LLMs for comparison, \ie, LLaMA-65B, MPT-7B, Dolly-12B, GPT-3, ChatGPT, and GPT-4, each in both the zero-shot and few-shot settings. A total of 14 models are included.
Note that out of the six LLMs, LLaMA-65 is the only one not trained with instruction-following tasks.
We use the crowd workers' annotation guidelines from CODA-19~\cite{huang2020coda} as our zero-shot prompt
(see \Cref{tab:prompt-zero-shot} in \Cref{appendix:prompt} for the actual prompt).
For the few-shot setting, we assume a scenario where users annotate a single abstract as an example.
Thus, we randomly select one abstract from CODA-19 train set 
that contains four primary labels (\textit{Background, Purpose, Method, and Finding}).
To avoid LLMs incorrectly considering the order of samples as information, 
we shuffle the samples in the few-shot prompt
(see \Cref{tab:prompt-few-shot} in \Cref{appendix:prompt} for the actual prompt).
To query each model, we use the parameters described in \Cref{appendix:hyperparameters}.
Once we obtain the generated texts, we use regex to parse the predicted label.
When the predicted label is not in the CODA-19 label sets or is missing, we treat it as \textit{Other}.
We report the per-label precision, recall, and F1 score, as well as the overall accuracy and macro F1 score.


\paragraph{Results.}
The results of our experiment are presented in \Cref{tab:exp-coda-19-no-data}.
SciBERT\textsubscript{CODA+Pos} remains the best-performing model with an accuracy of 0.698 and a macro F1 score of 0.709.
We observe that the zero-shot setting of LLaMA-65B performs poorly, possibly due to the model not being trained for any instruction-following tasks.
The majority of its predictions are on \textit{Background} and \textit{Other} labels, leading to very low recall for \textit{Purpose}, \textit{Method}, and \textit{Finding} labels (0.012, 0.004, and 0.056, respectively).
Such biased prediction issues also happen for MPT and Dolly in both zero-shot and few-shot settings even though they are trained with instruction-following datasets, suggesting that there is still a huge performance gap between open-sourced models and closed models.

When comparing closed models, our results show that ChatGPT performs worse than GPT-3, possibly due to its optimization toward human-favored conversation.
On the other hand, GPT-4 outperforms GPT-3 by a large margin but is still unable to outperform SciBERT\textsubscript{CODA+Pos}.
While we believe that LLMs have the potential to outperform SciBERT\textsubscript{CODA+Pos} in the future, our current results emphasize the importance of having a dedicated dataset that aligns well with the target task.

\section{Conclusion}

In this paper, we investigate the impact of different datasets and LLMs on the CODA-19 research aspect classification task.
Our findings show that using a huge but slightly different dataset, PubMed 200K, does not help improve performance.
LLMs, trained with massive web corpus, are also unable to outperform the SciBERT trained on the target dataset,
emphasizing the importance of task-aligned datasets.
In the future, we will explore methods for the model to consider the context and predict all sentences within a single abstract at once.

\section*{Acknowledgements}
We thank all anonymous reviewers' constructed feedback to improve this work.

\section*{Limitations}
One important aspect of achieving optimal performance when using LLMs is the design of a high-quality prompt.
In this study, we consider both zero-shot and few-shot learning scenarios, which assume no or very limited task-specific data.
However, iteratively refining the prompt over time to obtain the best-performing prompt may break the zero-shot or few-shot scenario.
Moreover, the final prompt used in this study is specifically designed to guide the crowd workers in the annotation process of the CODA-19 dataset, with frequently asked questions (FAQs) refined over time to address workers' confusion.
In a real-world scenario, users would not have access to such helpful FAQs when working on a new task.
Therefore, the performance of LLMs may be lower in practice.

Also, LLMs are susceptible to a data leakage problem due to their training with Internet data.
For example, ChatGPT is known to have been trained on Internet data prior to September 2021.
Considering that the CODA-19 dataset was released in July 2020, with its train, validation, and test sets made publicly available, there is a possibility that some closed models have seen the exact test instances, leading to an unfair comparison.
Since the training data are not disclosed for the closed models, the impact of this exposure on the models' performance remains unknown.

\section*{Ethics Statement}
Deploying the model in this paper would likely result in unknown false predictions. It requires further research to actually put it into practice.

\bibliography{bib/custom}
\bibliographystyle{acl_natbib}

\begin{table*}[t]
\small \centering
\begin{tabular}{@{}l@{\kern5pt}l@{\kern5pt}p{5.2in}@{}}
\toprule
    \textbf{Label} & \textbf{Id} & \textbf{Sentence} \\ \midrule
    Background & 0 & \textbf{[POSITION=0.00]} Breathing is a high-risk behavior for spreading infectious diseases in enclosed environments , \\
    Background & 1 & \textbf{[POSITION=0.07]} so it is important to investigate the characteristics of human exhalation flow and dispersal of exhaled air to reduce the risk . \\
    Background & 2 & \textbf{[POSITION=0.14]} This paper used two-dimensional time-resolved particle image velocimetry ( 2D TR-PIV ) to measure the exhaled flow from a breathing the rmal manikin . \\
    Method & 3 &  \textbf{[POSITION=0.21]} Since the exhaled flow is transient and periodic ,\\
    Method & 4 & \textbf{[POSITION=0.29]} the phase-averaged method was used to analyze the flow characteristics ., \\
\bottomrule
\end{tabular}
 \caption{A sample of position-encoded CODA-19 data extracted from the paper (\cite{feng2015tr}). Here, \textbf{Id} indicates the sentence index with respect to the abstract. Removing the position encoding (\eg, \textbf{[POSITION=0.00]}), we could get the original CODA-19 data.}
 \label{tab:data-coda}
\end{table*} 
\begin{table*}[t]
\small \centering
\begin{tabular}{@{}l@{\kern5pt}l@{\kern5pt}p{5.1in}@{}}
\toprule
    \textbf{Label} & \textbf{Id} & \textbf{Sentence} \\ \midrule
    Methods & 3 & \textbf{[POSITION=0.38]} The serum levels of follicle stimulating hormone (FSH), luteinizing hormone (LH), and estradiol (E(2)) were detected before and after the treatment. \\
    Results & 4 & \textbf{[POSITION=0.50]} After 12 weeks of treatment, HAMD scores in both groups decreased significantly (p<0.05) with no significant difference between the groups (p>0.05). \\
    Results & 5 & \textbf{[POSITION=0.62]} The levels of FSH decreased significantly and the level of E(2) increased significantly in both groups, and they changed more in the control group. \\
    Results & 6 & \textbf{[POSITION=0.75]} No side-effect of treatment was reported in either group during treatment. \\
    Conclusions & 7 & \textbf{[POSITION=0.88]} The Chinese medicinal formula GNL showed promise in relieving perimenopausal depression and merits further study. \\
\bottomrule
\end{tabular}
 \caption{A sample of position-encoded PubMed data extracted from Paper ID: 19769482. Here, \textbf{Id} indicates the sentence index with respect to the abstract. Removing the position encoding (\eg, \textbf{[POSITION=0.38]}), we could get the original PubMed data.}
 \label{tab:data-pubmed}
\end{table*} 

\newpage
\appendix

\section{Sample Data}
\label{appendix:sample-data}
In this section, we show some sample data for CODA-19 dataset (\Cref{tab:data-coda}) and the PubMed dataset (see \Cref{tab:data-pubmed}). As shown in the table, [POSITION=0.38] is the position encoding we added to inject the positional information.

\section{Training and Testing Details}
\label{appendix:hyperparameters}
Here, we describe all the training details for the models we build in this study.
All the models are trained using PyTorch~\cite{NEURIPS2019_9015} and HuggingFace~\cite{wolf2019huggingface}.

\begin{itemize}[leftmargin=*,itemsep=0mm]
    \item
    \textbf{SciBERT\textsubscript{CODA}.}
    We fine-tune SciBERT using the original CODA-19 training set using the hyperparameters listed in \Cref{tab:hyperparameters}.

    \item
    \textbf{SciBERT\textsubscript{CODA+Pos}.}
    The position encoding is first added to create the position-encoded CODA-19 dataset.
    We then fine-tune SciBERT on the created dataset using the hyperparameters listed in \Cref{tab:hyperparameters}.

    \item
    \textbf{SciBERT\textsubscript{PubMed}.}
    Similar to the above but on the original PubMed dataset.

    \item
    \textbf{SciBERT\textsubscript{PubMed+Pos}.}
    Similar to the above but on the position-encoded PubMed dataset.

    \item
    \textbf{SciBERT\textsubscript{Mix+Pos+S}.}
    We first turn the position-encoded PubMed's label into the CODA-19 label space using the pre-defined mapping: \textit{Background} to \textit{Background}, \textit{Methods} to \textit{Method}, \textit{Objective} to \textit{Purpose}, \textit{Results} to \textit{Finding}, and \textit{Conclusions} to \textit{Finding}.
    Second, we simply mix the position-encoded CODA-19 training set with the position-encoded PubMed training set together to create a simply-mixing dataset.
    We then fine-tune SciBERT on the created dataset using the hyperparameters listed in \Cref{tab:hyperparameters}.
    Note that the CODA-19 validation set is used to checkpoint the best model.

    \item
    \textbf{SciBERT\textsubscript{Mix+Pos+U}.}
    As the data sizes of CODA-19 and PubMed differ a lot (137K vs. 2.2M in the training set), 
    we upsample position-encoded CODA-19's training set 10 times to create a more balanced upsampling dataset.
    Before mixing, PubMed's label has been transferred to the CODA-19 label space using the pre-defined mapping function.
    We then fine-tune SciBERT using this dataset with the pre-defined hyperparameters (\Cref{tab:hyperparameters}).
    Again, the CODA-19 validation set is used to checkpoint the best model.

    \item
    \textbf{SciBERT\textsubscript{Mix+Pos+T}.}
    For the two-staged training strategy using PubMed and CODA-19 dataset,
    we first fine-tune SciBERT on the position-encoded PubMed dataset with the pre-defined hyperparameters (\Cref{tab:hyperparameters}).
    Here, the PubMed validation set is used to checkpoint the best model.
    In the second stage, we fine-tune the checkpointed model on the position-encoded CODA-19 dataset with the pre-defined hyperparameters (\Cref{tab:hyperparameters}).

    \item
    \textbf{LLaMA.}
    We obtain LLaMA-65B from the official Github\footnote{\url{https://github.com/facebookresearch/llama}} and run the generation using HuggingFace's interface~\cite{wolf2019huggingface}.
    Temperature sampling is used for text generation with temperature = 0.1, num\_beams = 1, top\_p = 0.95, repetition\_penalty = 1.0, min\_new\_tokens = 1, and max\_new\_tokens = 10.

    \item
    \textbf{MPT.}
    We use \texttt{mosaicml/mpt-7b-instruct} and run the generation using HuggingFace's interface~\cite{wolf2019huggingface} with the same parameters described in LLaMA.
    
    \item
    \textbf{Dolly.}
    We use \texttt{databricks/dolly-v2-12b} and run the generation using HuggingFace's interface~\cite{wolf2019huggingface} with the same parameters described in LLaMA.
    

    \item
    \textbf{GPT-3.}
    We use \texttt{text-davinci-003} with the parameters: temperature = 0.0, max\_tokens = 10, top\_p = 0.95, frequency\_penalty = 0.0, and presence\_penalty = 0.0.
    

    \item
    \textbf{ChatGPT.}
    We use \texttt{gpt-3.5-turbo} with the same parameters described in GPT-3.
    Note that when calling ChatGPT, we simply put all the prompts in a single user input.

    \item
    \textbf{GPT-4.}
    We use \texttt{gpt-4} with the same parameters described in GPT-3.
    Again, when calling GPT-4, we simply put all the prompts in a single user input.
    
\end{itemize}

\begin{table}[t]
    \centering \small
    \begin{tabular}{ll} 
        \toprule
        \textbf{Hyperparameter} & \textbf{Value} \\ \midrule
        \textbf{Model} & scibert\_scivocab\_uncased \\ 
        \textbf{Batch Size} & 64 \\
        \textbf{Learning Rate} & 1e-5 \\
        \textbf{Epochs} & 20 \\
        \textbf{Metric for Best Model} & Evaluation Accuracy \\ 
        \textbf{Max Sequence Length} & 128 \\ 
        \textbf{Warmup Ratio} & 0.1 \\ 
        \textbf{Early Stopping Patience} & 6 \\
        \bottomrule
    \end{tabular}
     \caption{General hyperparameters used for training the models. We used HuggingFace's Trainer for fine-tuning all the models. Parameters not specified here remain the default values.}
     \label{tab:hyperparameters}
\end{table}

\section{Prompts}
\label{appendix:prompt}

\Cref{tab:prompt-zero-shot} and \Cref{tab:prompt-few-shot} show the zero-shot prompt and the few-shot prompt we used for querying LLMs. 


\begin{table*}
    \centering \small
    \begin{tabular}{@{}p{16cm}@{}}
    \toprule
    \textbf{Zero-shot Prompt} \\
    Classify the given text into one of the following labels. \\
     \\

    [Background]: Text segments answer one or more of these questions: Why is this problem important?, What relevant works have been created before?, What is still missing in the previous works?, What are the high-level research questions?, How might this help other research or researchers? \\ 
    
    [Purpose]: Text segments answer one or more of these questions: What specific things do the researchers want to do?, What specific knowledge do the researchers want to gain?, What specific hypothesis do the researchers want to test?  \\ 
    
    [Method]: Text segments answer one or more of these questions: How did the researchers do the work or find what they sought?, What are the procedures and steps of the research?  \\ 
    
    [Finding]: Text segments answer one or more of these questions: What did the researchers find out?, Did the proposed methods work?, Did the thing behave as the researchers expected?  \\ 
    
    [Other]: Text fragments that do NOT fit into any of the four categories above. Text fragments that are NOT part of the article. Text fragments that are NOT in English. Text fragments that contains ONLY reference marks (e.g., "[1,2,3,4,5") or ONLY dates (e.g., "April 20, 2008"). Captions for figures and tables (e.g. "Figure 1: Experimental Result of ...", or "Table 1: The Typical Symptoms of ...") Formatting errors. I really don't know or I'm not sure.  \\ 
    
     \\
    FAQs \\
    1. This text fragment has terms that I don't understand. What should I do? Please use the context in the article to figure out the focus. You can look up terms you don't know if you feel like you need to understand them.  \\
    2. This text fragment is too short to mean anything. What should I do? If the text fragment is too short to have significant meanings, you could consider the entire sentence and answer based on the entire sentence.  \\
    3. This text fragment is NOT in English. What should I do? If the whole fragment (or the majority of words in the fragment) is in Non-English, please label it as "Other". If the majority of the words in this fragment are in English with a few non-English words, please judge the label normally.  \\
    4. I'm not sure if this should be a "background" or a "finding." How do I tell? When a sentence occurs in the earlier part of an article, and it is presented as a known fact or looks authoritative, it is often a "background" information.  \\
    5. Do "potential applications of the proposed work" count as "background" or "purpose"? It should be "background." The "purpose" refers to specific things the paper wants to achieve.  \\
    6. If the article says it's a "literature review" (e.g., "We reviewed the literature" / "In this article, we review.." etc), would we classify those as finding/contribution or purpose? Most parts of a literature review paper should still be "background" or "purpose", and only the "insight" drew from a set of prior works can be viewed as a "finding/contribution".  \\
    7. What should I do with the case study on a patient? Typically, it has a patient come in with a set of signs and symptoms in the ER, and then the patient gets assessed and diagnosed. The patient is admitted to the hospital ICU and tests are done and they may be diagnosed with something else. In such cases, please label the interventions done by the medical staff (e.g., CT scans, X-rays, and medications given) as "Method", and the patient's final result (e.g. the patient's pneumonia resolved and he was released from the hospital) as "Finding/Contribution".  \\
     \\
    Classify the following sentence into one of the label: Background, Purpose, Method, Finding, and Other.  \\
    \\ 
    
    Text: \textbf{```\{Target-Sentence\}'''}  \\
    The answer label for Text is [ \\
    \bottomrule
    \end{tabular}
    \caption{Zero-shot prompt used when calling LLMs (LLaMA, MPT, Dolly, GPT-3, ChatGPT, and GPT-4). The \textbf{\{Target-Sentence\}} will be replaced by the sentence we would like to predict.}
    \label{tab:prompt-zero-shot}
    
\end{table*}

\begin{table*}
    \centering \small
    \begin{tabular}{@{}p{16cm}@{}}
    \toprule
    \textbf{Few-shot Prompt} \\
    Classify the given text into one of the following labels. \\
     \\

    [Background]: Text segments answer one or more of these questions: Why is this problem important?, What relevant works have \\
    
    $\cdots$ \\
    
    $\cdots$ (Same as the zero-shot prompt that describe the definition of the label and FAQs. Skip for space.) \\

    $\cdots$ \\

    Classify the following sentence into one of the label: Background, Purpose, Method, Finding, and Other.  \\
    \\ 

    Text: ```With the features of extremely high selectivity and efficiency in catalyzing almost all the chemical reactions in cells ,'''  \\
    Label: [Background]  \\
    
    Text: ```enzymes play vitally important roles for the life of an organism and hence have become frequent targets for drug design .'''  \\
    Label: [Background]  \\
    
    Text: ```by which users can easily obtain their desired results .'''  \\
    Label: [Method]  \\
    
    Text: ```An essential step in developing drugs by targeting enzymes is to identify drug-enzyme interactions in cells .'''  \\
    Label: [Background]  \\
    
    Text: ```a user-friendly web server was established ,'''  \\
    Label: [Method]  \\
    
    Text: ```called `` iEzy-Drug , '' in which each drug compound was formulated by a molecular fingerprint with 258 feature components ,'''  \\
    Label: [Method]  \\
    
    Text: ```and the prediction engine was operated by the fuzzy K-nearest neighbor algorithm .'''  \\
    Label: [Method]  \\
    
    Text: ```Although some computational methods were developed in this regard based on the knowledge of the three-dimensional structure of enzyme ,'''  \\
    Label: [Background]  \\
    
    Text: ```Here , we reported a sequence-based predictor ,'''  \\
    Label: [Purpose]  \\
    
    Text: ```Moreover , to maximize the convenience for the majority of experimental scientists ,'''  \\
    Label: [Method]  \\
    
    Text: ```It is both time-consuming and costly to do this purely by means of experimental techniques alone .'''  \\
    Label: [Background]  \\
    
    Text: ```The overall success rate achieved by iEzy-Drug via rigorous cross-validations was about 91 \% .'''  \\
    Label: [Finding]  \\
    
    Text: ```unfortunately their usage is quite limited because threedimensional structures of many enzymes are still unknown .'''  \\
    Label: [Background] \\
    
    Text: ```each enzyme by the Chou 's pseudo amino acid composition generated via incorporating sequential evolution information and physicochemical features derived from its sequence ,'''   \\    
    Label: [Method]  \\
    
    Text: \textbf{```\{Target-Sentence\}'''}  \\
    The answer label for Text is [ \\
    
    \bottomrule
    \end{tabular}
    \caption{Few-shot prompt used when calling LLMs (LLaMA, MPT, Dolly, GPT-3, ChatGPT, and GPT-4). The \textbf{\{Target-Sentence\}} will be replaced by the sentence we would like to predict. The skipped label description and FAQs are the same as the zero-shot prompt (see \Cref{tab:prompt-zero-shot}). The few-shot samples are from one single abstract to simulate the scenario where people annotate some data as a reference.}
    \label{tab:prompt-few-shot}
\end{table*}

\end{document}